%% file: Main.tex
\documentclass[preprint]{elsarticle}
\usepackage{xpatch}
\usepackage{array}
\newcolumntype{P}[1]{>{\centering\arraybackslash}m{#1}}
\usepackage[table]{xcolor} 

\usepackage{booktabs} 
\usepackage{adjustbox}
\usepackage{multirow}
\usepackage{amsmath,amssymb,amsfonts}
\usepackage{algorithm}
\usepackage{algpseudocode}
\usepackage{graphicx}
\usepackage{textcomp}
\usepackage{xcolor}
\usepackage[flushleft]{threeparttable}
\usepackage{enumitem}
\usepackage[bottom]{footmisc}
\usepackage{longtable}
\usepackage{caption}
\usepackage{subcaption} 
\usepackage{afterpage}
\usepackage[normalem]{ulem}
\usepackage{comment}

\usepackage{url}
\usepackage[hidelinks]{hyperref}
\xpatchcmd{\State}{\algorithmicend\ \algorithmicfor}{\algorithmicend}{}{}

\usepackage{float} 

\algtext*{EndFor}
\algtext*{EndIf}

\begin{document}

\begin{frontmatter}

\title{Adaptive Unknown Fault Detection and Few-Shot Continual Learning for Condition Monitoring in Ultrasonic Metal Welding}
\tnotetext[preprintnote]{Submitted to \textit{Machine Learning: Engineering} (IOP). This is a preprint version.}




\author[UIUCMechSE,UIUCCSL]{Ahmadreza Eslaminia}
\ead{ae15@illinois.edu}

\author[UIUCMechSE]{Kuan-Chieh Lu}
\ead{kclu3@illinois.edu}

\author[UIUCCSL]{Klara Nahrstedt}
\ead{klara@illinois.edu}

\author[UMich,UIUCMechSE]{Chenhui Shao\corref{mycorrespondingauthor}}
\ead{chshao@umich.edu}

\address[UIUCMechSE]{Department of Mechanical Science and Engineering, University of Illinois at Urbana-Champaign, Urbana, IL 61801, USA}
\address[UIUCCSL]{Coordinated Science Laboratory, University of Illinois at Urbana-Champaign, Urbana, IL 61801, USA}
\address[UMich]{Department of Mechanical Engineering, University of Michigan, Ann Arbor, MI 48109, USA}

\cortext[mycorrespondingauthor]{Corresponding author}

\begin{abstract}

Ultrasonic metal welding (UMW) is widely used in industrial applications but is sensitive to tool wear, surface contamination, and material variability, which can lead to unexpected process faults and unsatisfactory weld quality. Conventional monitoring systems typically rely on supervised learning models that assume all fault types are known in advance, limiting their ability to handle previously unseen process faults. To address this challenge, this paper proposes an adaptive condition monitoring approach that enables unknown fault detection and few-shot continual learning for UMW. Unknown faults are detected by analyzing hidden-layer representations of a multilayer perceptron and leveraging a statistical thresholding strategy. Once detected, the samples from unknown fault types are incorporated into the existing model through a continual learning procedure that selectively updates only the final layers of the network, which enables the model to recognize new fault types while preserving knowledge of existing classes. To accelerate the labeling process, cosine similarity transformation combined with a clustering algorithm groups similar unknown samples, thereby reducing manual labeling effort. Experimental results using a multi-sensor UMW dataset demonstrate that the proposed method achieves 96\% accuracy in detecting unseen fault conditions while maintaining reliable classification of known classes. After incorporating a new fault type using only five labeled samples, the updated model achieves 98\% testing classification accuracy. These results demonstrate that the proposed approach enables adaptive monitoring with minimal retraining cost and time. The proposed approach provides a scalable solution for continual learning in condition monitoring where new process conditions may constantly emerge over time and is extensible to other manufacturing processes.

\end{abstract}


\end{frontmatter}


\input{introduction}

\input{methodology}

\input{result}

\input{conclusion}

\section*{Acknowledgments}
This research has been supported by the National Science Foundation, United States under Grant No. 2433484.

\bibliographystyle{elsarticle-num-names} 

\bibliography{mybibfile}

\end{document}

%% file: introduction.tex
\section{Introduction} \label{sec:introduction}

Ultrasonic metal welding (UMW) is a solid-state joining process with unique advantages compared to conventional fusion-based welding methods including the ability to join similar or dissimilar materials, short process cycles, and environmental friendliness \cite{martinsen2015joining,cai2017ultrasonic}. As such, UMW has widespread industrial applications such as electric vehicle (EV) battery pack assembly \cite{Kim2011,banerjee2025quantifying},  automotive body joining \cite{haddadi2016grain,Ni2018,becker2021multi}, and hybrid metal-polymer heat exchangers \cite{meng2019ultrasonic,kuntumalla2019joining}.

Despite these advantages, UMW has a relatively narrow operating window and is susceptible to uncontrollable process disturbances~\cite{nong2018improving,lu2026integrated}, such as tool degradation, material surface contamination, and material variability~\cite{Lee2014,nazir2021online,lu2026sensor}. In industrial environments, such disturbances are inevitable and often lead to inconsistent weld quality, increased scrap rates, and the need for frequent manual intervention~\cite{cai2017ultrasonic,shao2013feature}. Consequently, online monitoring has been extensively studied to detect process anomalies and predict weld quality \cite{feng2024review,geng2025machine}.

A large body of research has focused on improving monitoring performance in UMW through sensor fusion, feature engineering, and machine learning model development. Early studies primarily adopted statistical process control methods for quality classification in EV battery assembly. For example, \citet{shao2013feature} proposed a cross-validation-based method to simultaneously select features and tune monitoring limits. \citet{Lee2014} developed a physics-based approach by correlating online power and displacement signals with weld attributes such as bond density and postweld thickness. \citet{Guo2016} combined univariate control charts with Mahalanobis distance to construct flexible decision boundaries, which achieved near-zero misdetection rates. Subsequent work explored machine learning and deep learning techniques for improved monitoring accuracy \cite{balz2020process,schwarz2022improving,Meng2022}. Another line of research has focused on predicting joint strength through solving a regression problem \cite{zhao2017effect,Wu2022,Meng2022}. In addition to weld quality prediction, tool condition monitoring (TCM) \cite{shao2016tool,nazir2021online,dong2025fine} and surface condition monitoring have been extensively studied \cite{lu2023online,tian2023weldmon,lu2026sensor}. Recent efforts have incorporated domain generalization \cite{meng2024meta,eslaminia2024federated}, domain adaptation, and explainable few-shot learning \cite{Meng2023} to improve robustness under varying operating conditions.

Despite these advances, existing UMW monitoring methods face several critical challenges in real-world implementations. Most current approaches assume that all fault types or process conditions are known during training~\cite{geng2021open}. In practice, however, new fault types may emerge due to unexpected disturbances, tool degradation, or changes in material properties \cite{chen2020robust}. As a result, models trained on predefined classes often fail to recognize previously unseen conditions, leading to incorrect predictions. This highlights a fundamental limitation of existing methods in detecting unknown process faults.

Addressing this limitation is further complicated by significant data-related challenges in UMW monitoring. Data-driven models typically require a large amount of high-quality labeled data, the collection of which is costly, labor-intensive, and time-consuming in manufacturing environments \cite{naghavi2025large, eslaminia2024federated}. Moreover, sensing data in UMW are high-dimensional and heterogeneous, making it difficult to learn robust and generalizable representations, especially when only limited samples are available \cite{meng2024meta}.

Even when unknown faults can be detected, incorporating them into the monitoring model is a major challenge. Retraining the entire model is computationally expensive and may lead to catastrophic forgetting~\cite{kirkpatrick2017overcoming,boschini2022class}, while newly observed fault types usually have only a small number of labeled samples~\cite{zhao2023few,Meng2023}. In addition, manual labeling of unknown fault data is time-consuming and can significantly delay model updates, creating a critical bottleneck for practical deployment.

In the broader machine learning literature, unknown fault detection and continual learning have been separately studied to address similar challenges. Existing approaches for unknown fault detection often rely on out-of-distribution detection, distance-based methods, or confidence-based thresholding to identify samples that deviate from known classes, e.g., \cite{theissler2017detecting,chen2020robust,wang2024unknown}. Continual learning methods aim to incrementally update models while preserving previously learned knowledge, using strategies such as regularization, replay, or parameter isolation \cite{boschini2022class,zhou2024class}. However, these methods cannot be directly applied to UMW monitoring. First, industrial monitoring data are typically high-dimensional, noisy, and strongly influenced by process-specific characteristics, making generic detection methods less effective. Second, many continual learning approaches require large amounts of labeled data or extensive retraining, which is impractical in manufacturing environments where new fault data are scarce~\cite{eslaminia2025fdm,tian2025machinestethoscope}. Third, most existing methods do not consider the labeling bottleneck, which is critical for enabling rapid adaptation in real-world systems.

To address these challenges, this paper develops an adaptive approach for condition monitoring in UMW that integrates unknown fault detection with few-shot continual learning. Unknown faults are identified by analyzing hidden-layer representations of a multilayer perceptron (MLP) using a statistical thresholding strategy. Once detected, new fault types are incorporated into the model through a selective update procedure that retrains only the final layers, enabling efficient adaptation while preserving previously learned knowledge. To further improve practicality, a cosine similarity-based transformation combined with BIRCH (balanced iterative reducing and clustering using hierarchies) clustering is used to group similar unknown samples and reduce labeling effort. Experimental results demonstrate that the proposed method achieves 96\% accuracy in detecting unseen fault conditions, with all unknown samples correctly identified. Moreover, new fault types can be incorporated using as few as five labeled samples, achieving up to 98\% overall accuracy while maintaining strong classification performance for existing classes.

To the best of the authors' knowledge, this paper is the first to address unknown fault detection and few-shot continual learning for condition monitoring in UMW. The main contributions of this work are summarized as follows:
\begin{enumerate}[label=(\arabic*)]
    \item \textbf{Unknown fault detection:} A data-driven method that leverages hidden-layer representations of a multilayer perceptron (MLP) combined with class-specific statistical thresholding to reliably identify previously unseen fault conditions without modifying the trained model.
    
    \item \textbf{Few-shot continual learning:} A selective model update strategy that incorporates new fault types using only a small number of labeled samples by retraining only the final layers, thereby preserving previously learned knowledge while enabling efficient adaptation under data-scarce conditions.
    
    \item \textbf{Efficient labeling via clustering:} A similarity-based data transformation using cosine similarity, coupled with BIRCH clustering, to group unknown samples and significantly reduce manual labeling effort, enabling rapid model updating in practical deployment.
\end{enumerate}

The remainder of this paper is organized as follows. Section~\ref{sec:method} presents the proposed methods for unknown fault detection, continual learning, and clustering. Section~\ref{sec:result} describes the experimental setup and evaluates the performance of the proposed approach. Section~\ref{sec:conclusion} concludes the paper and suggests future research directions.

%% file: methodology.tex
\section{Methodology}\label{sec:method}
The proposed adaptive monitoring approach integrates three components: (i) an unknown fault detection strategy that identifies samples deviating from the distributions of known classes by analyzing hidden-layer representations of an MLP, (ii) a few-shot continual learning procedure that updates the classifier to incorporate newly labeled fault types while preserving previously learned knowledge, and (iii) a clustering-based labeling mechanism that groups unknown samples in a similarity space to reduce manual labeling effort. Together, these components enable efficient fault detection, rapid model adaptation, and practical deployment under evolving process conditions.

\begin{figure}[ht]
    \centering
    \includegraphics[width=0.99\textwidth]{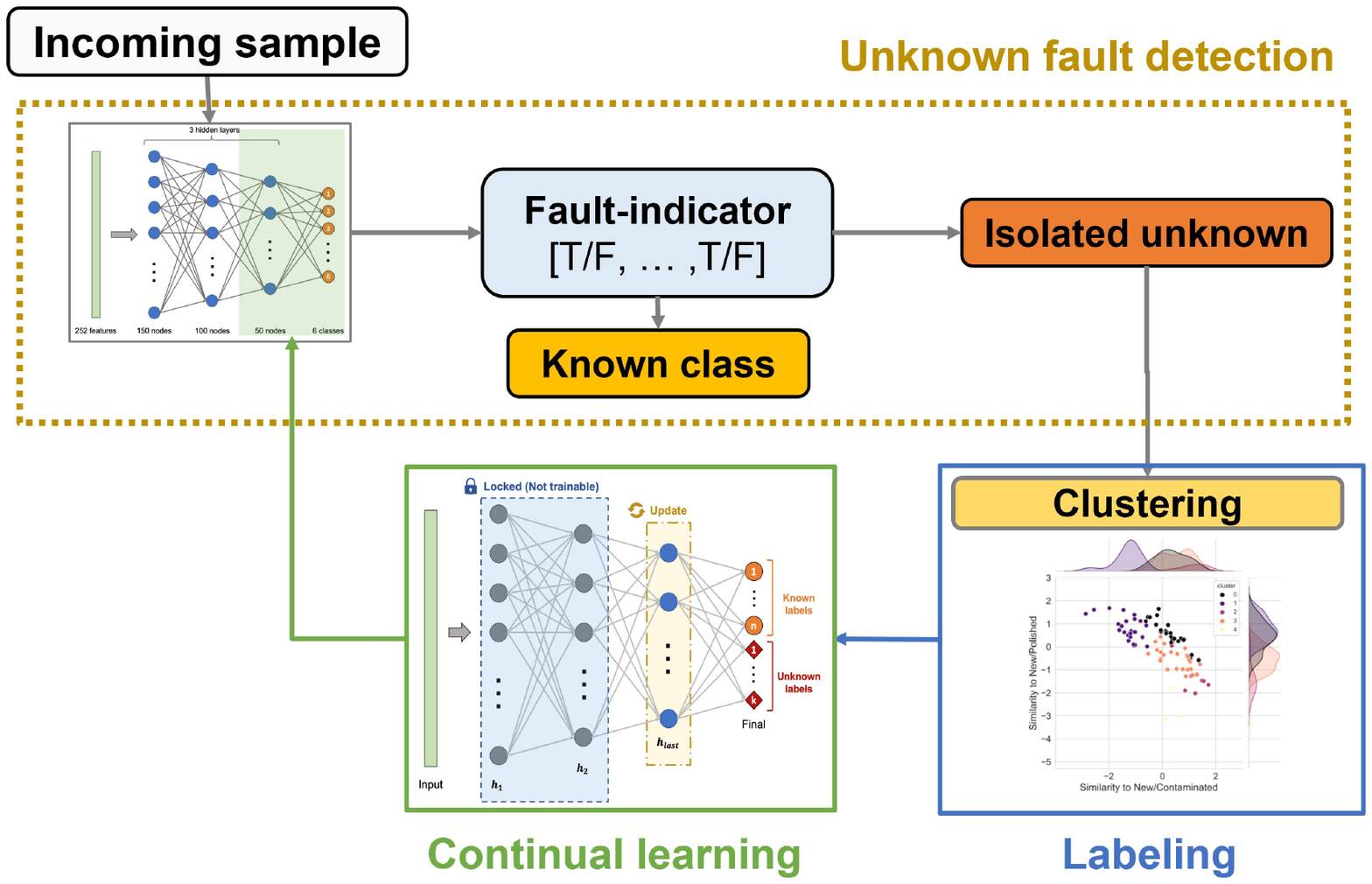}
    \caption{Overview of the proposed adaptive monitoring framework, consisting of unknown fault detection, cluster-assisted labeling, and few-shot continual learning.}
    \label{fig:overview}
\end{figure}

Fig.~\ref{fig:overview} illustrates the interaction among the three components within the proposed approach. Incoming samples are first processed by the unknown fault detection module, which either assigns them to known fault classes or flags them as potential unknown faults. The flagged samples are subsequently grouped using clustering to reduce labeling effort. The newly labeled data are then used to selectively update the model, enabling the incorporation of new fault types while preserving previously learned knowledge. Sections~\ref{subsec:unknown_fault}--\ref{sub_sec:clustering} provide detailed descriptions of these three components.

\subsection{Unknown fault detection}\label{subsec:unknown_fault}

\begin{figure}[htb]
    \centering
    \includegraphics[width=0.99\textwidth]{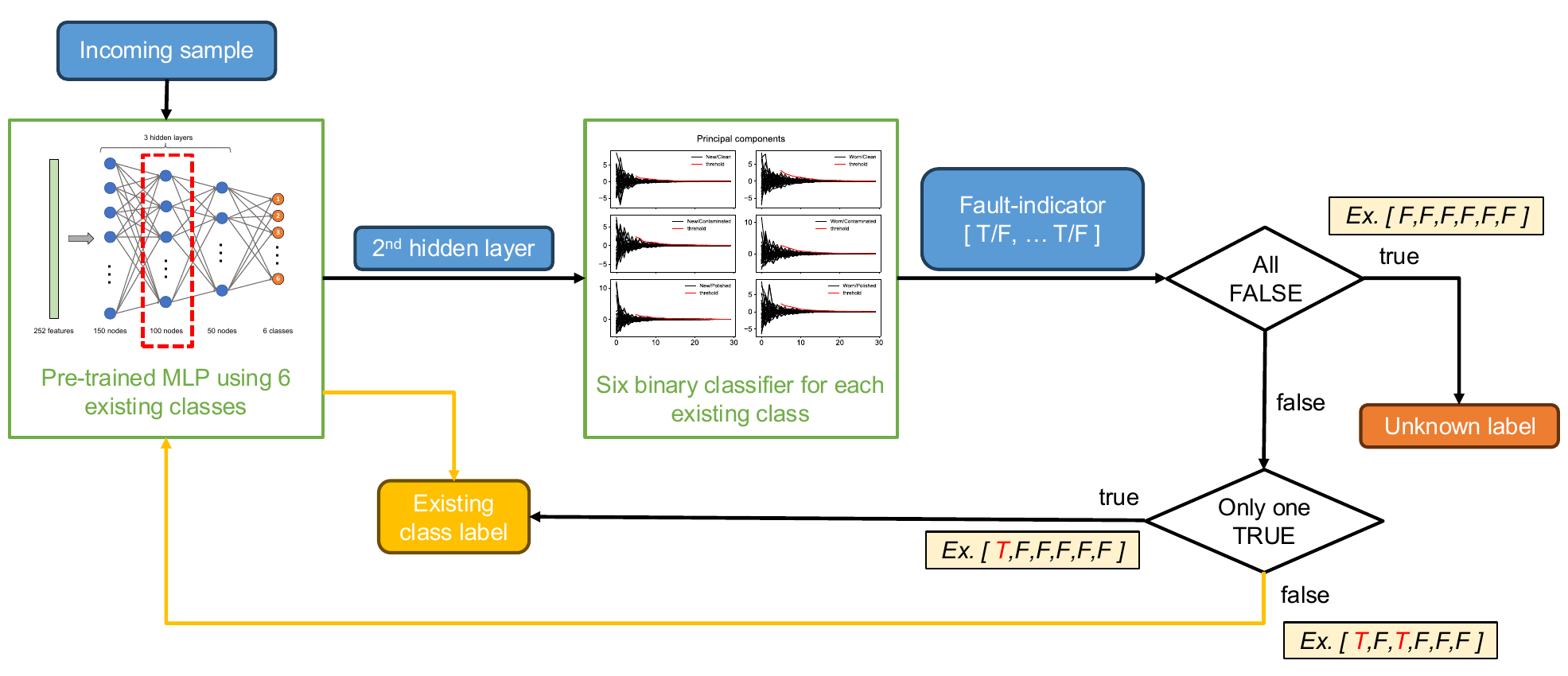}
    \caption{Schematic for the proposed unknown fault detection method}
    \label{UFD_framework}
\end{figure}

Traditional classification models rely on the final output layer to assign a sample to a predefined class, which limits their ability to detect fault types that were not included in the training data. To overcome this limitation, the proposed framework leverages the rich feature representations embedded in the hidden layers of a trained MLP, which preserve essential physical information throughout the feature transformation process. These intermediate representations retain sufficient distributional information to distinguish previously unseen fault types from known classes. Fig.~\ref{UFD_framework} illustrates the unknown fault detection method, and Algorithm~\ref{alg:ufd} summarizes the complete procedure in detail.

The following provides the mathematical formulation of the key steps in Algorithm~\ref{alg:ufd}. Let $x\in\mathbb{R}^{d}$ denote an input sample and consider an MLP with parameters $\theta$. The hidden-layer mapping is
\begin{equation}
h^{(\ell)}(x)=\phi\!\left(W^{(\ell)}h^{(\ell-1)}(x)+b^{(\ell)}\right),\qquad h^{(0)}(x)=x,
\end{equation}
where $\phi(\cdot)$ is the ReLU activation. An intermediate embedding is then extracted as
\begin{equation}
z(x)=h^{(\ell)}(x)\in\mathbb{R}^{q},
\end{equation}
where the layer index $\ell$ is treated as a hyperparameter of the framework.

\begin{algorithm}[H]
\caption{Unknown Fault Detection Method}
\label{alg:ufd}
\begin{algorithmic}[1]
\small
\Require Training data $\mathcal{D}=\{(x_i,y_i)\}_{i=1}^{N}$, known classes $\mathcal{C}=\{1,\dots,C\}$,
         hidden layer index $\ell$, number of principal component analysis (PCA) components $r$
\Ensure  Fault-indicator $\mathbf{I}(x)$ and class assignment for each incoming sample $x$

\Statex \textbf{--- Offline Training Phase ---}
\State Train MLP $f_{\theta}$ on $\mathcal{D}$; fix all parameters $\theta$
\For{each known class $c \in \mathcal{C}$}
    \State Extract embeddings: $\mathcal{Z}_c = \{h^{(\ell)}(x) : (x,y)\in\mathcal{D},\; y=c\}$
    \State Standardize $\mathcal{Z}_c$ (zero mean, unit variance) $\;\rightarrow\; \tilde{\mathcal{Z}}_c$
    \State Fit PCA on $\tilde{\mathcal{Z}}_c$; store principal directions $P_c \in \mathbb{R}^{q \times r}$
    \State Compute training PCA scores: $U_c = \tilde{\mathcal{Z}}_c\, P_c$
    \For{$p = 1,\dots,r$}
        \State Set threshold: $t_{c,p} = 3\cdot\mathrm{std}(U_{c,p})$
    \EndFor
\EndFor

\Statex \textbf{--- Online Detection Phase ---}
\For{each incoming sample $x$}
    \State Extract embedding: $z(x) = h^{(\ell)}(x)$
    \For{each known class $c \in \mathcal{C}$}
        \State Standardize $z(x)$ using class-$c$ statistics $\;\rightarrow\; \tilde{z}_c(x)$
        \State Compute PCA score: $u_c(x) = P_c^{\top}\tilde{z}_c(x)$
        \State Compute consistency indicator: $I_c(x)=\displaystyle\prod_{p=1}^{r}\mathbb{I}\!\left(|u_{c,p}(x)|\leq t_{c,p}\right)$
    \EndFor
    \State Construct fault-indicator: $\mathbf{I}(x)=[I_1(x),\dots,I_C(x)]$
    \If{$\displaystyle\sum_{c=1}^{C} I_c(x) = 0$}
        \State \textbf{Flag} $x$ as an unknown fault \hfill$\triangleright$ \textit{Proceed to clustering and labeling}
    \ElsIf{$\displaystyle\sum_{c=1}^{C} I_c(x) = 1$}
        \State \textbf{Assign} $x$ to class $c^{*}$ where $I_{c^*}(x)=1$
    \Else
        \State \textbf{Assign} $x$ to $c^{*}=\arg\max_{c}\,[f_{\theta}(x)]_c$ \hfill$\triangleright$ \textit{Use MLP softmax output}
    \EndIf
\EndFor
\end{algorithmic}
\end{algorithm}

For each known class $c\in\{1,\dots,C\}$, PCA is applied to the embeddings of that class. Let $\tilde{z}_c(x)$ denote the standardized embedding and let the columns of $P_c$ be the principal directions learned from class $c$. The PCA score vector is
\begin{equation}
u_c(x)=P_c^{\top}\tilde{z}_c(x)\in\mathbb{R}^{r}.
\end{equation}
A class-specific acceptance region is specified by component-wise bounds
\begin{equation}
|u_{c,p}(x)| \le t_{c,p}, \qquad p=1,\dots,r,
\end{equation}
where the bounds are determined using a three-sigma rule derived from the statistical distribution of PCA scores for known samples in class $c$. The class-consistency indicator is then defined as
\begin{equation}
I_c(x)=\prod_{p=1}^{r}\mathbb{I}\!\left(|u_{c,p}(x)| \le t_{c,p}\right),
\end{equation}
and the corresponding fault-indicator vector is
\begin{equation}
\mathbf{I}(x)=\left[I_1(x),\dots,I_C(x)\right].
\end{equation}

Once deployed, all incoming samples are processed through the framework, producing a fault-indicator vector $\mathbf{I}(x)$. The interpretation of this vector leads to three possible outcomes:
\begin{enumerate} [label=Case \arabic*:, leftmargin=*]
    \item \text{All elements are FALSE:} The sample cannot be associated with any known class and is flagged as an unknown fault for further investigation.
    \item \text{Exactly one element is TRUE:} The sample is assigned to the corresponding known fault class.
    \item \text{More than one element is TRUE:} The sample is consistent with multiple known classes. To resolve this ambiguity, the sample is passed through the full MLP, and the final softmax output is used to assign the most probable class label.
\end{enumerate}

By leveraging hidden-layer representations and statistical thresholds, the framework enables rapid and reliable classification while maintaining adaptability to evolving fault conditions. Since unknown fault detection is performed without modifying the trained model parameters, the approach is well suited for real-world applications where new fault types may emerge over time.

\subsection{Continual learning} \label{subsec:CL}
\begin{figure}[htb]
    \centering
    \includegraphics[width=0.7\textwidth]{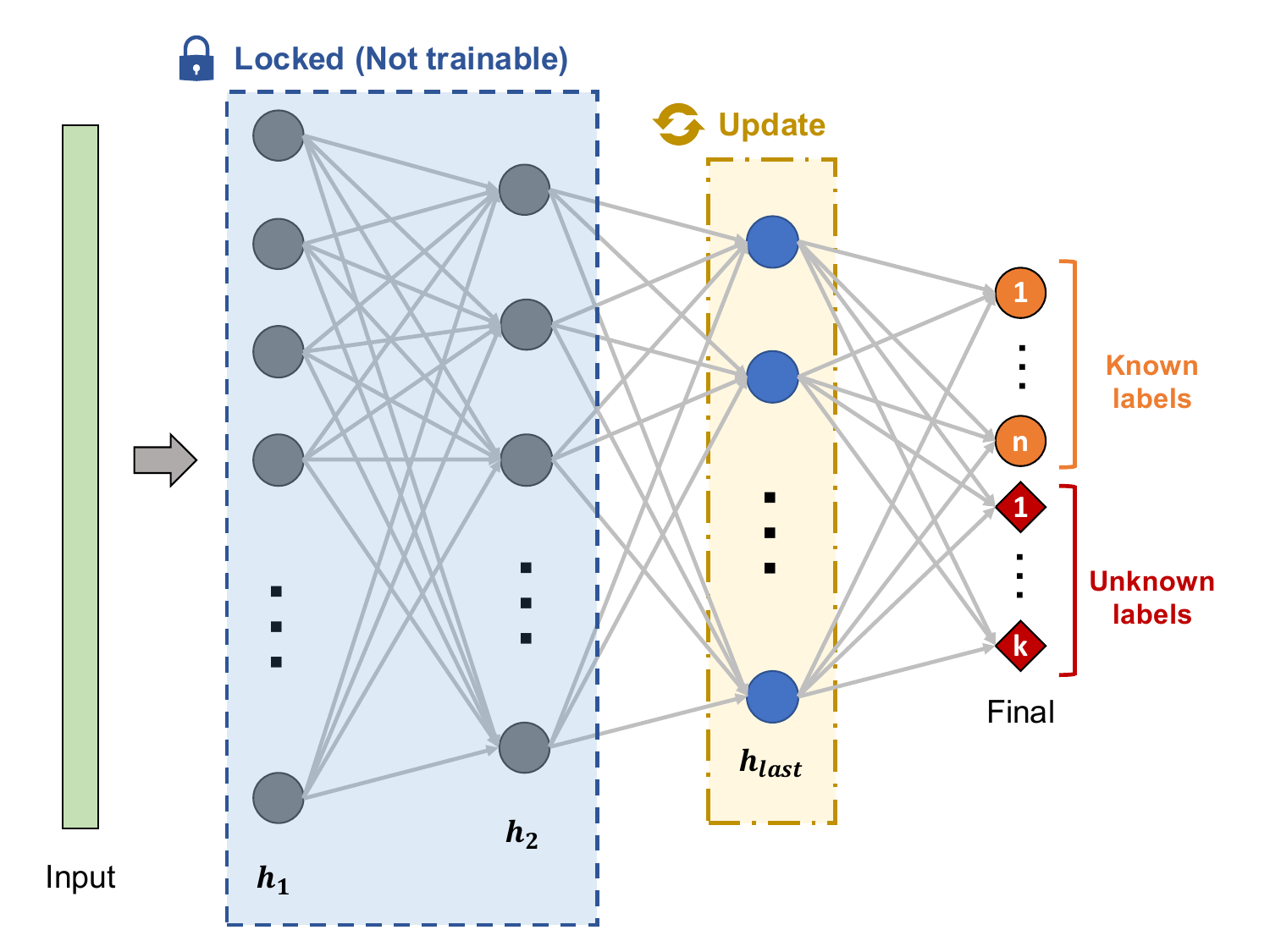}
    \caption{Model updating process of continual learning}
    \label{fig:model_update}
\end{figure}

Once an unknown fault has been detected and labeled, the condition monitoring model must be updated to accommodate the new fault type while preserving its ability to classify existing classes. Fig.~\ref{fig:model_update} illustrates the proposed model update procedure. Starting from a pretrained MLP designed for classifying $C$ known fault types, the output layer is extended by adding $k$ new nodes, where each additional node corresponds to a newly discovered fault class.

Assume an initial classifier is trained on $C$ known classes with parameters $\theta$. When $k$ new fault types are discovered, the classifier is expanded to predict over $C+k$ classes. We partition the parameters into frozen layers $\theta_f$ and updatable layers $\theta_u$. Model updating is performed by optimizing
\begin{equation}
\min_{\theta_u}\; \sum_{(x,y)\in\mathcal{D}_{\mathrm{update}}} -\log p_{\theta}(y\,|\,x)
\quad \text{s.t.}\quad \theta_f\ \text{fixed},
\end{equation}
where $\mathcal{D}_{\mathrm{update}}$ denotes the labeled data available for adaptation and $p_{\theta}(y\,|\,x)$ is the softmax probability of the expanded network.

The key idea is to freeze the early hidden layers, whose weights $\theta_f$ have already been optimized to extract discriminative fault features, and retrain only the later layers $\theta_u$ (including the expanded output layer) to accommodate the new classification space. By preserving the learned feature representations in the frozen layers, previously acquired knowledge remains intact, while the retrained layers adapt to the expanded set of fault classes. Since multiple output nodes can be added simultaneously, this procedure enables the model to incorporate more than one unknown fault type at a time.

From a computational perspective, this selective update strategy substantially reduces the cost of model retraining compared to optimizing all parameters from scratch. By freezing the early layers, the number of trainable parameters is significantly reduced, which not only improves computational efficiency but also enables effective model updates using a small number of samples from newly introduced fault types. Modifying the early layers is not only unnecessary but also risks destabilizing the feature space that underpins classification of existing classes. By restricting updates to $\theta_u$, the adaptation process remains efficient and stable, making the approach suitable for real-time monitoring and large-scale fault detection systems. This strategy also facilitates incremental learning, allowing the model to evolve dynamically as new fault conditions arise without requiring full retraining.

Since newly introduced fault types typically have far fewer labeled samples than existing classes, the limited data representation may pose challenges during model adaptation, particularly when multiple unknown fault types must be incorporated simultaneously. Nonetheless, the selective update mechanism effectively balances adaptability with knowledge preservation, enabling efficient model expansion even under data-scarce conditions.

\subsection{Clustering} \label{sub_sec:clustering}

With an accurate unknown fault detection system and an efficient model update mechanism in place, a remaining challenge lies in the labeling process for newly identified unknown faults. Updating the model requires assigning appropriate class labels to all flagged unknown samples, which in practice involves careful examination of each sample to determine its corresponding root cause (class label). Manually investigating every sample is time-consuming and may substantially delay the model update process. To address this bottleneck, clustering can significantly reduce the required labeling effort by grouping similar samples together, enabling operators to label an entire cluster by inspecting only a few representative points.

Effective clustering requires a data transformation that maps samples into a space where their separability is enhanced. Since data from known fault classes are available, measuring the similarity between an unknown sample and the known classes provides a natural basis for distinguishing different fault types. Cosine similarity is adopted as the distance metric for this purpose. The cosine similarity between two vectors $A$ and $B$ is defined as:
\begin{equation}
    \text{Cosine similarity}\quad S(A,B)= \frac{A \cdot B}{\|A\| \|B\|},
\end{equation}
\noindent where $A \cdot B$ denotes the dot product and $\|A\|\|B\|$ is the product of their magnitudes; higher values indicate greater directional alignment between the two vectors.

Each unknown sample is transformed into a compact similarity-based representation with respect to the known classes. Let $z(x)$ denote the hidden-layer embedding of sample $x$. For each known class $i$, an average cosine similarity is computed between embeddings of sample $x$ and all samples in that class:
\begin{equation}
s_i(x)=\frac{1}{|\mathcal{D}_i|}\sum_{x' \in \mathcal{D}_i}
\frac{z(x)\cdot z(x')}{\|z(x)\|\,\|z(x')\|},
\end{equation}
where $\mathcal{D}_i$ is the set of known samples from class $i$. This yields a similarity vector
\begin{equation}
s(x) = [s_1(x), s_2(x), \dots, s_C(x)] \in \mathbb{R}^{C},
\end{equation}
whose $i$th entry quantifies the similarity of the unknown sample to known class $i$. A clustering algorithm is then applied to the set of similarity vectors $\{s(x)\}$ to group unknown samples into clusters. Samples within the same cluster share similar relationships to the known classes, enabling efficient labeling by inspecting only a small number of representative samples per cluster.

Cosine similarity is adopted as the distance metric due to its suitability for high-dimensional fault data \cite{xia2015learning,mehta2023greedy,meng2024meta}. Unlike Euclidean distance \cite{jain1999data}, which measures absolute differences and loses discriminatory power in high-dimensional spaces, cosine similarity evaluates the directional alignment between vectors, making it scale-invariant and robust to feature magnitude variations. Beyond its discriminative properties, this transformation maps each sample to a low-dimensional similarity vector whose entries carry a clear physical meaning as ``similarity to a known condition,'' reducing the effective dimensionality of the clustering input and improving interpretability.

BIRCH~\cite{zhang1997birch} is selected as the clustering algorithm due to its ability to automatically determine the number of clusters, process data in a single scan, and support incremental updates without reprocessing the entire dataset---properties that are well suited for continual learning scenarios where new fault types emerge over time.

%% file: result.tex
\section{Results and Discussion}\label{sec:result}
\subsection{Experimental setup}

\begin{figure}[ht]
    \centering
    \includegraphics[width=\textwidth]{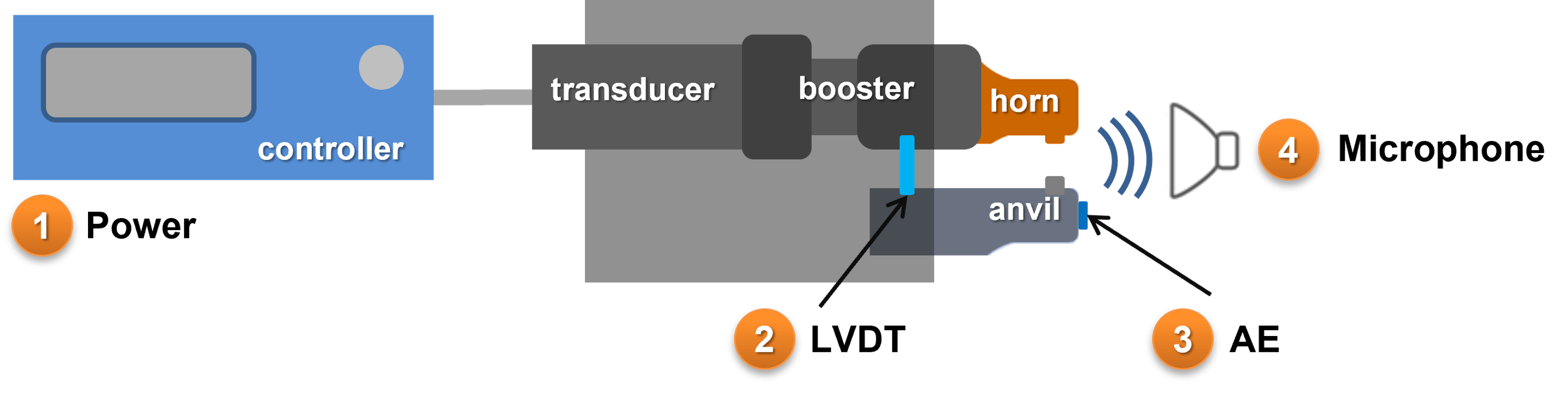}
    \caption{Online monitoring system for UMW.}
    \label{fig:monitoring_system}
\end{figure}

\begin{figure}[ht]
    \centering
    \includegraphics[width=0.5\textwidth]{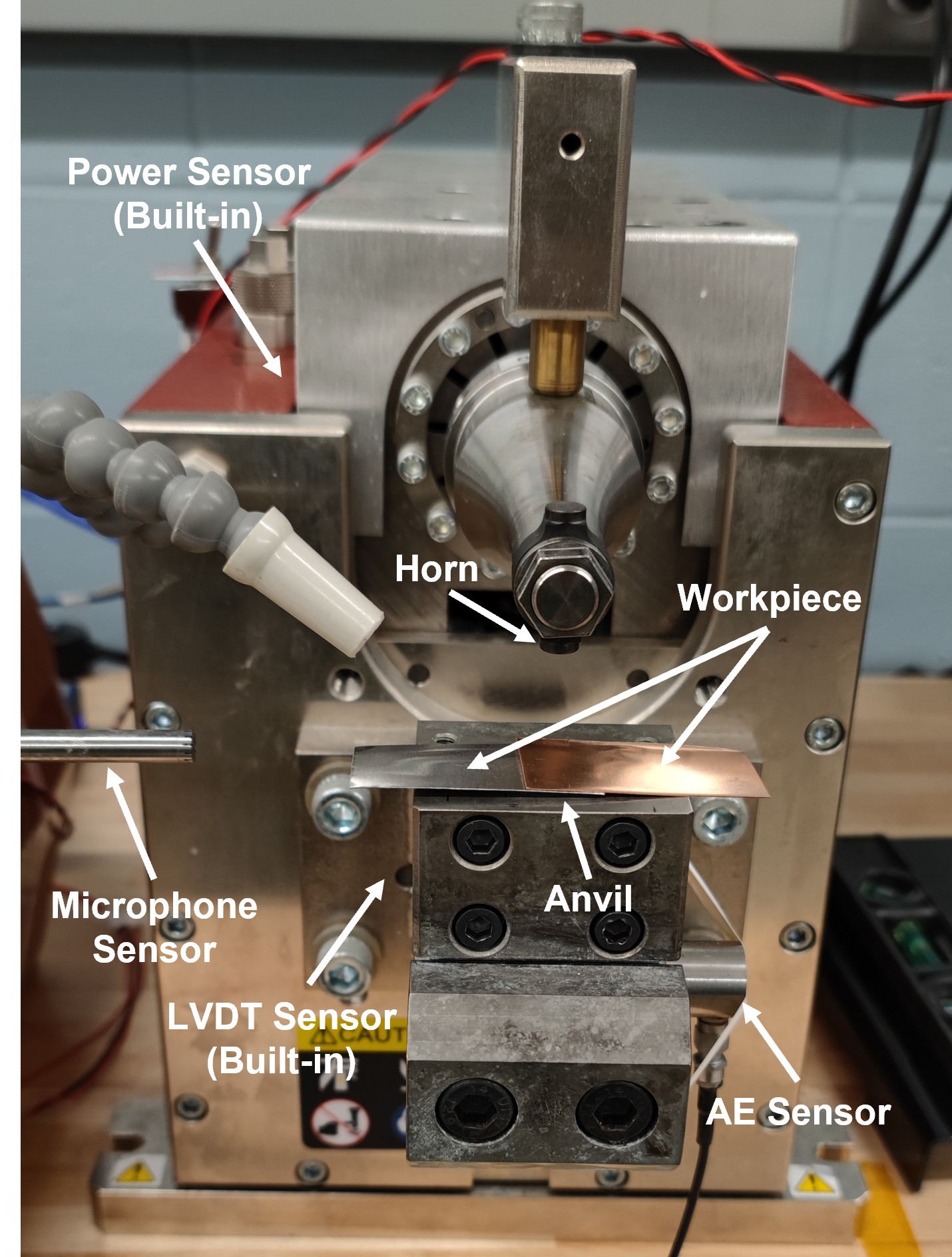}
    \caption{Photo of the UMW machine with sensors \cite{meng2024meta}.}
    \label{fig:UMW_photo}
\end{figure}

The experiments in this study utilize a multi-sensor UMW dataset collected using the Branson Ultraweld L20 ultrasonic welding machine, which is equipped with an online monitoring system comprising four sensors: a built-in linear variable differential transformer (LVDT), a built-in power sensor, an acoustic emission (AE) sensor, and a microphone~\cite{lu2026sensor}. The monitoring system is illustrated by Figs. \ref{fig:monitoring_system} and \ref{fig:UMW_photo}.

The dataset covers mixed process conditions involving three tool conditions (new, worn, and damaged) and three surface conditions (clean, contaminated, and polished), yielding nine classes with 30 samples each (270 samples in total). Combinations of the new and worn tool conditions with all three surface types form six known fault classes used for initial model training, while all damaged tool conditions are withheld during training and treated as unknown faults for evaluation. For predictor construction, we adopt the feature set identified by the genetic-algorithm-based cost- and time-efficient sensor and feature selection framework developed in~\cite{lu2026sensor}, which jointly optimizes classification accuracy, sensor cost, and feature extraction latency. This study focuses on unknown fault detection, continual learning, and clustering. For continual learning, we study both single-class and multi-class updates under a data-scarce setting in which each known class has approximately 25 samples and each newly introduced class has at most 5 labeled samples.

Across all studies, we use an MLP with three fully connected hidden layers of sizes $150$, $100$, and $50$, with ReLU activations, followed by a softmax classification layer. Model optimization is performed using the Adam optimizer with categorical cross-entropy loss. For known-class model evaluation, we use stratified $5$-fold cross-validation to preserve class balance across folds.

\subsection{Unknown fault detection}
The unknown fault detection method is evaluated with respect to two key outcomes: (1) unknown fault types should be accurately identified and flagged for further investigation, and (2) known fault types should be correctly classified without excessive false alarms.

In this study, the hidden representation is extracted from the second hidden layer of the trained MLP (the $100$-dimensional embedding after the ReLU activation). For each known class, a class-specific standardization (z-score normalization) is applied, followed by PCA in the standardized embedding space. A three-sigma bound is computed for each retained PCA score, and a test sample is flagged as unknown when it fails the acceptance test for all known classes.

\begin{figure}[ht]
    \centering
    \includegraphics[width=0.99\textwidth]{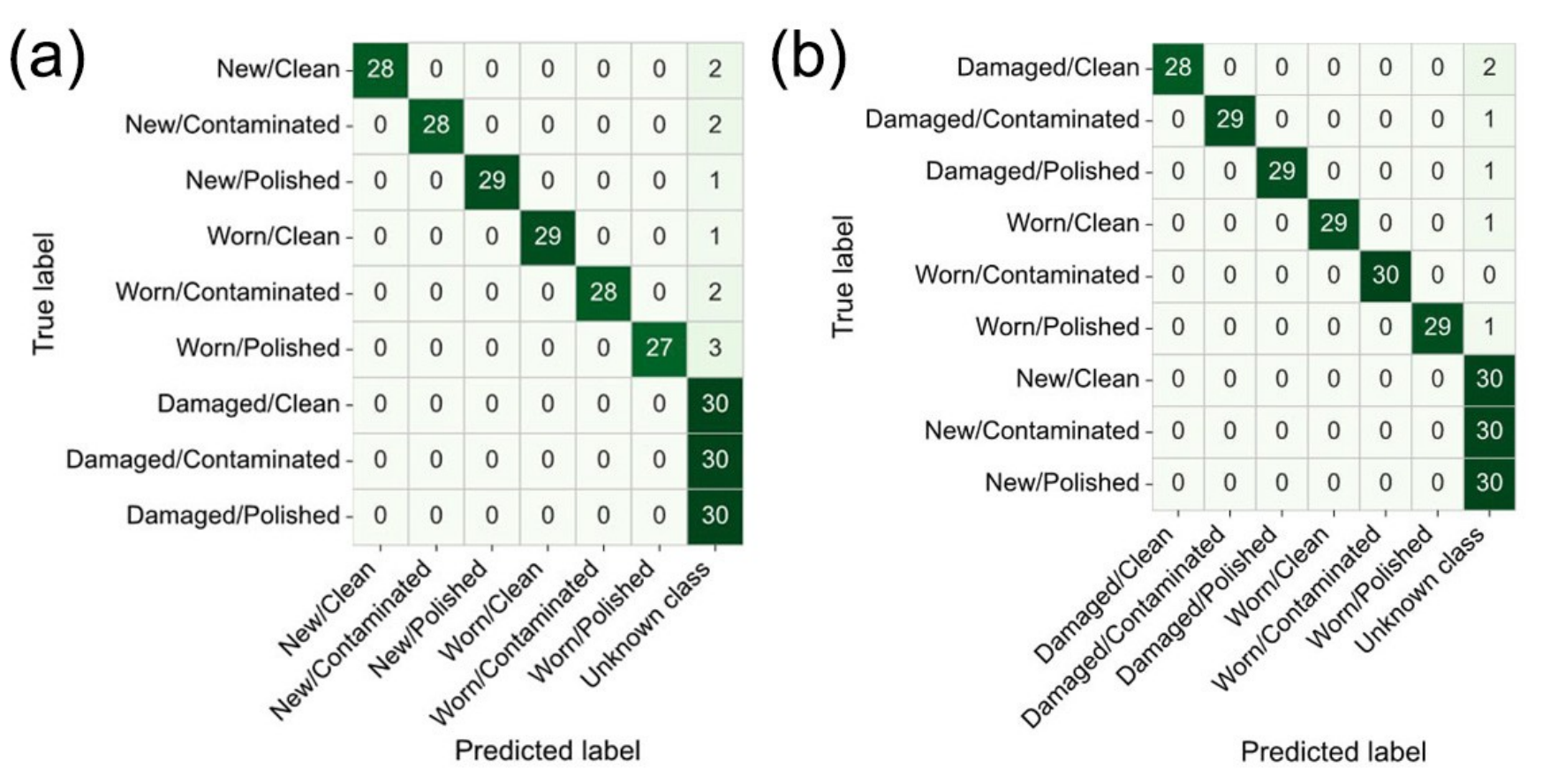}
    \caption{Unknown fault detection results: (a) damaged tool and (b) new tool.}
    \label{unknown}
\end{figure}

To evaluate performance, we first consider a scenario where a damaged tool is treated as the unknown fault type. Unlike typical tool wear, which follows a gradual degradation pattern, unexpected events such as tool collision or breakage can introduce distinct failure modes that must be detected early to prevent process disruptions. As shown in Fig.~\ref{unknown}a, the proposed method achieves 96\% overall detection accuracy, with 100\% unknown recall and 94\% classification accuracy on known classes, accompanied by minimal false alarms.

The method also demonstrates strong robustness under more complex conditions. In the damaged tool dataset, three distinct surface conditions are present, increasing task difficulty. Despite this, the proposed approach consistently identifies unknown faults without performance degradation. Additional experiments with alternative unknown classes in mixed-condition datasets (Fig.~\ref{unknown}b) further confirm that the method maintains reliable detection performance across different fault scenarios while preserving classification accuracy for known classes.

Moreover, most samples fall into two outcome categories---either all elements of the fault-indicator are FALSE or only one element is TRUE---indicating that the method enables efficient one-step classification. Once the fault-indicator is generated, the system can directly assign a known class label or flag the sample as unknown without additional decision-making, eliminating the need for hierarchical classification.

\subsection{Continual learning} \label{subsec:continual_learning}
This case study evaluates whether the proposed continual learning procedure can incorporate new fault types while maintaining high classification accuracy for existing classes. Two scenarios are considered: the introduction of a single unknown fault type, and the simultaneous addition of multiple unknown fault types. The six known classes (combinations of new and worn tool conditions with three surface types) are used for initial training, while all damaged tool conditions are withheld and designated as unknown faults for evaluation.

An additional challenge is imposed through data imbalance: each known class contains approximately 25 data points, whereas the newly introduced unknown classes have at most 5 labeled samples. This setup reflects real-world conditions where newly discovered fault types have limited available data.

The pretrained MLP weights are used to initialize the model, and the softmax layer is expanded to include additional output nodes for the new fault type(s). The first two hidden layers ($h_1$ and $h_2$) are frozen, and the remaining trainable layers ($h_{\text{last}}$ and the output layer) are fine-tuned using Adam with categorical cross-entropy under the same training protocol described in the experimental setup.

\begin{figure}
    \centering
    \includegraphics[width=0.8\textwidth]{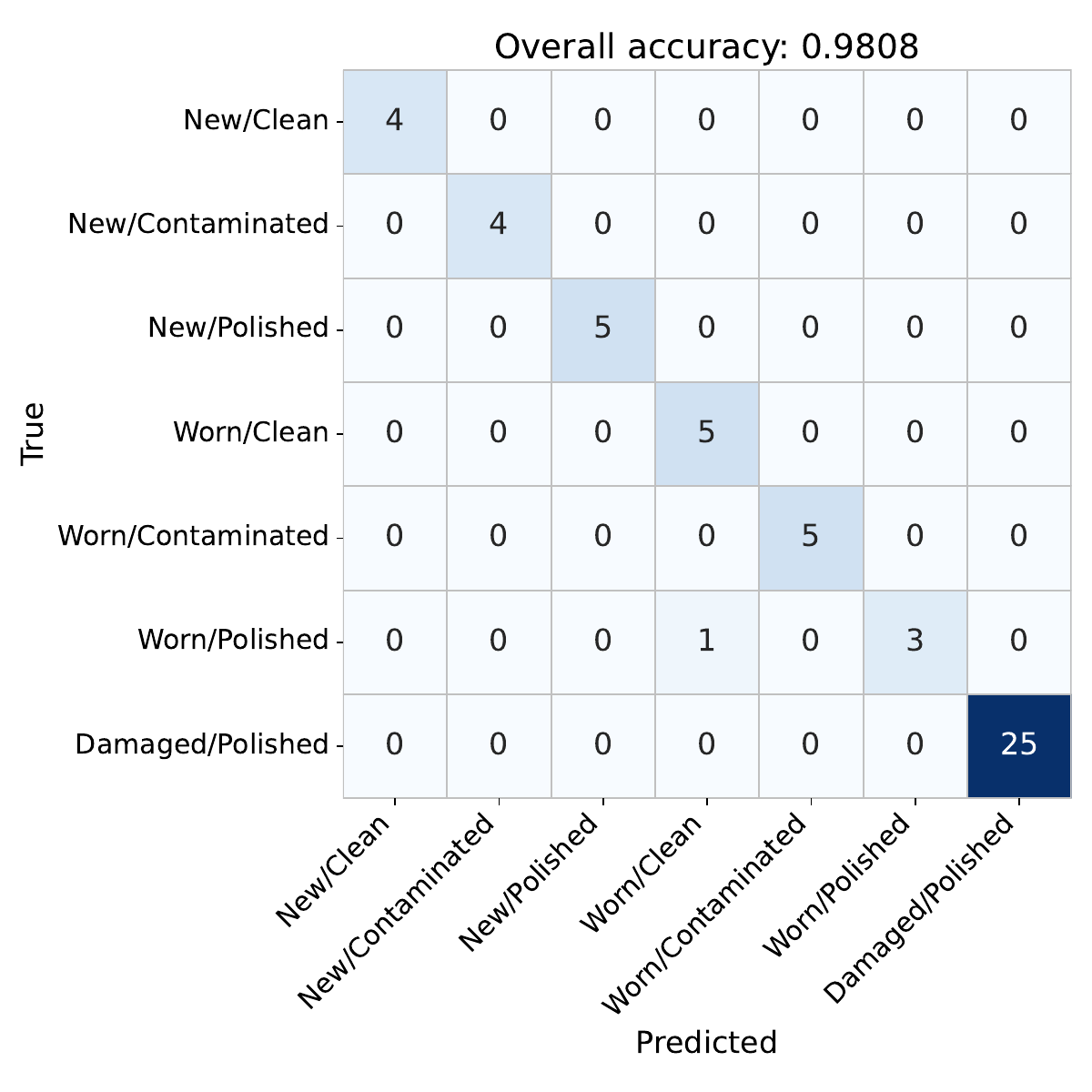}
    \caption{Confusion matrix after incorporating one new fault class using five labeled samples}
    \label{1c5new}
\end{figure}

Fig.~\ref{1c5new} shows the confusion matrix when a single unknown fault type is introduced. All data points in the confusion matrix represent unseen samples that were not part of the training process, ensuring that the evaluation reflects the model's generalization ability. With only five labeled samples for the new class, the updated model achieves 98\% accuracy, with all misclassifications occurring within the known fault classes rather than mislabeling the unknown fault type.

\begin{figure}
    \centering
    \includegraphics[width=\textwidth]{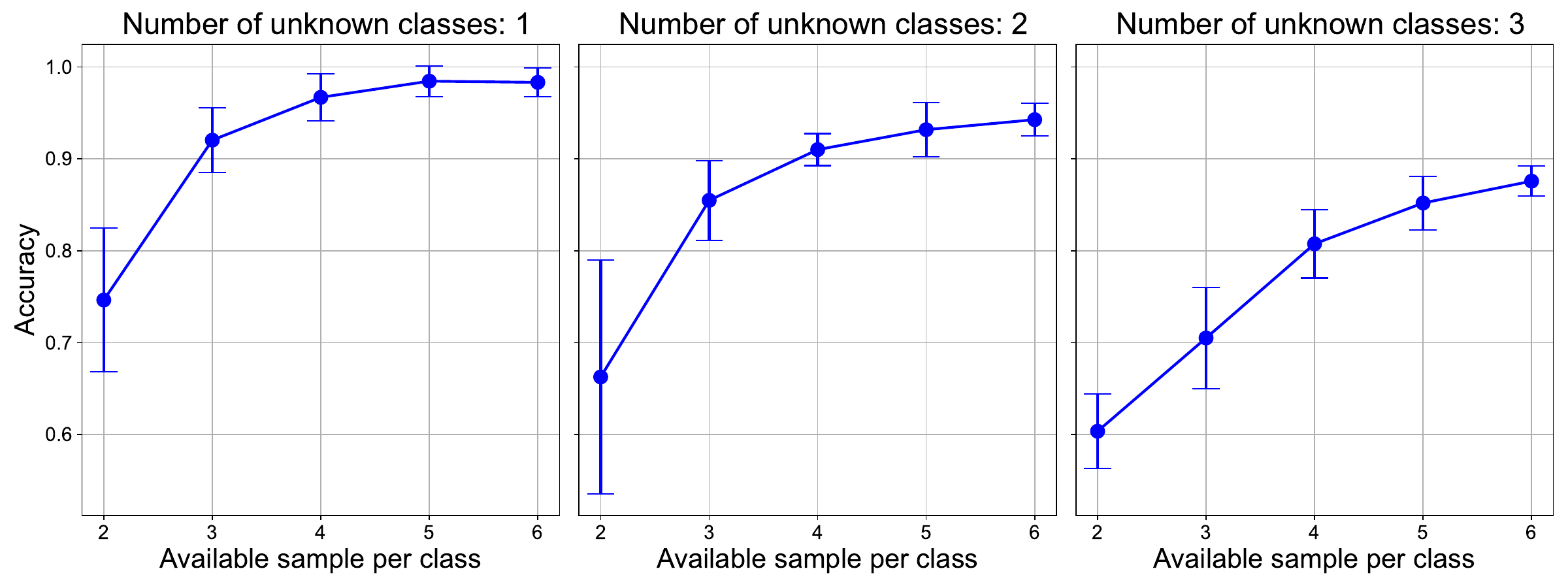}
    \caption{Classification accuracy as a function of the number of unknown classes and the number of labeled samples per class. Error bars denote one standard deviation over 20 repeated trials, reflecting the variability of model performance under different data sampling conditions.}
    \label{CL_stat}
\end{figure}

In the second case study, multiple unknown fault types are introduced simultaneously to increase task complexity, while the number of available samples per fault is intentionally limited to simulate severe data scarcity. Specifically, the number of new fault classes is varied from 1 to 3, and the number of samples per class ranges from 2 to 6. Each configuration is repeated 20 times to evaluate performance robustness, and the results are summarized in Fig.~\ref{CL_stat}.

When only a single unknown fault is introduced, the model consistently achieves high accuracy even with very few samples. However, as task complexity increases (e.g., incorporating three unknown fault types simultaneously), performance degrades under extremely limited data conditions, particularly in distinguishing between unknown classes. In addition, the standard deviation of classification accuracy increases as the number of samples decreases, indicating greater sensitivity of the update process to the quality and representativeness of the available data.

\subsection{Clustering}

In this case study, the feature representation for each sample is extracted from the second hidden layer of the MLP, yielding a $100$-dimensional embedding in $\mathbb{R}^{100}$. The cosine-similarity transformation maps each sample to a vector of average similarities to the six known classes, producing a six-dimensional representation in $\mathbb{R}^{6}$. For each sample, the similarity vector is computed by averaging its cosine similarity to all samples within each known class, and the resulting vectors are standardized before clustering. BIRCH is applied with automatic cluster count selection and a threshold parameter of $2.0$.

\begin{figure}[H]
    \centering
    \includegraphics[width=\textwidth]{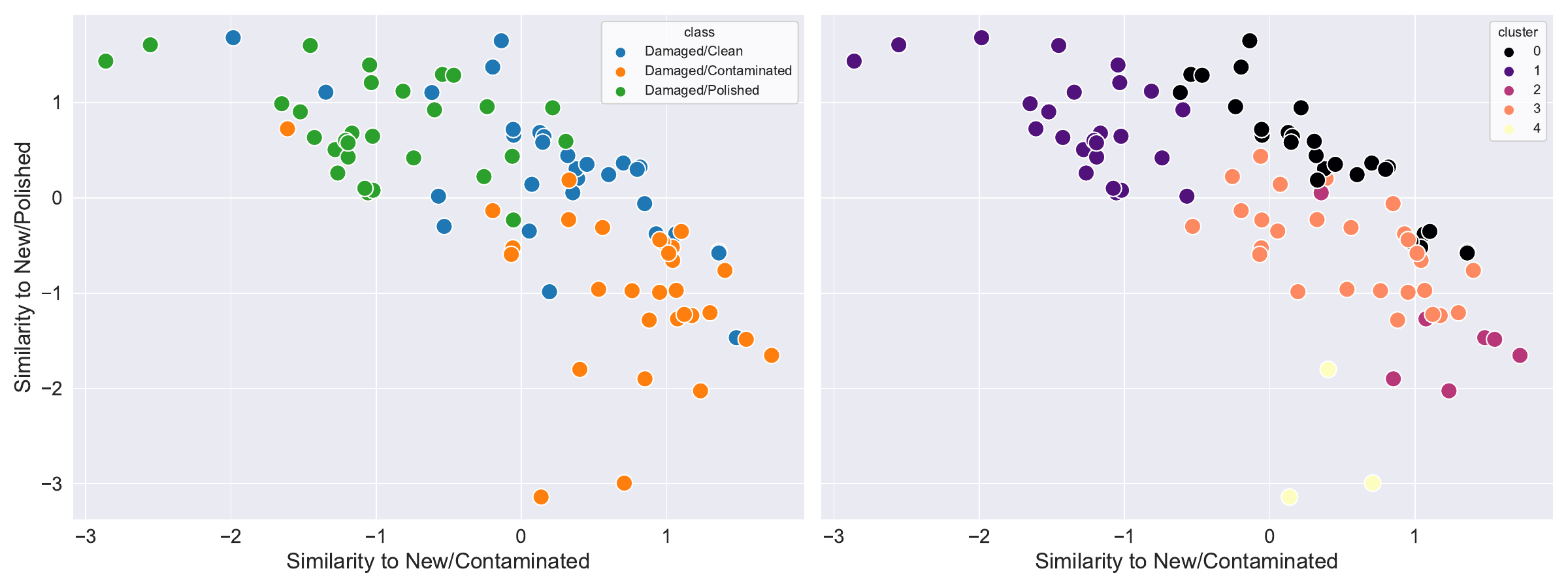}
    \caption{BIRCH clustering results}
    \label{cluster}
\end{figure}

\begin{figure}[H]
    \centering
    \includegraphics[width=0.7\textwidth]{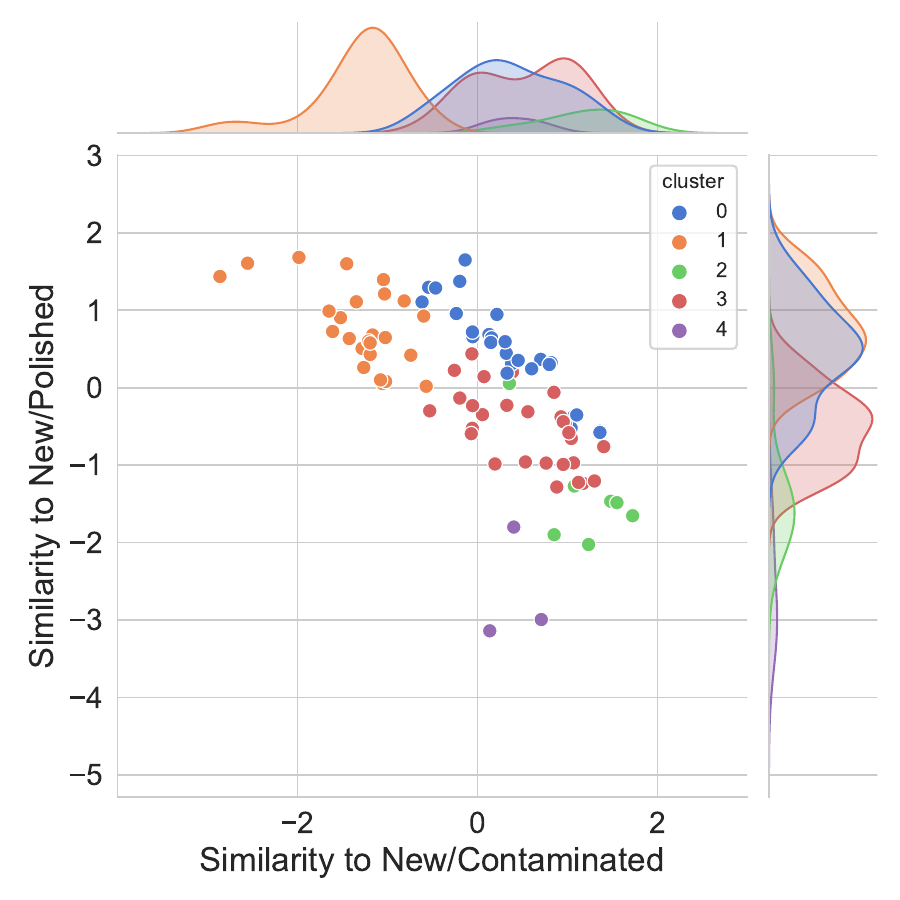}
    \caption{Clustering result in the selected cosine similarity space}
    \label{cluster_dist}
\end{figure}

Fig.~\ref{cluster} shows the BIRCH clustering results, where the dataset is grouped into five clusters with an overall clustering purity of 72\%. While not perfect, these results provide an efficient starting point for fault investigation in practice. The method rapidly groups data points that exhibit clear separability while reserving time and resources for more ambiguous cases. Fig.~\ref{cluster_dist} reveals a fuzzy region where samples from clean surfaces exhibit a wider distribution and overlap with one another, making them inherently difficult to distinguish, even for human inspectors.

Despite these challenges, the proposed method effectively groups the easily distinguishable samples, significantly reducing the manual effort required for fault analysis. As demonstrated in Section \ref{subsec:continual_learning}, the continual learning procedure requires only a small number of well-labeled samples to incorporate new fault types. Therefore, even if the clustering results are not perfectly accurate, obtaining a small subset of correctly labeled points from the preliminary clustering is sufficient to initiate the model update process. A more detailed labeling process can be conducted subsequently to refine the dataset for future use.

%% file: conclusion.tex
\section{Conclusion}\label{sec:conclusion}

This paper presents an adaptive approach for condition monitoring in UMW that enables unknown fault detection, novel class labeling, and few-shot continual learning. The proposed method addresses key challenges in industrial UMW, including the presence of previously unseen fault types, limited availability of labeled data, and the need for efficient model updates without retraining from scratch. Experimental results demonstrate that the proposed method achieves 96\% accuracy in detecting unseen fault classes, with all unknown samples correctly identified. Furthermore, new fault types can be incorporated using as few as five labeled samples, achieving up to 98\% overall accuracy while maintaining strong classification performance for existing classes. These results demonstrate the effectiveness of the proposed approach in both fault detection and model adaptation under data-scarce conditions. In addition, the incorporation of clustering-based labeling significantly reduces manual effort by grouping similar unknown samples, enabling faster and more efficient model updates. This capability enhances the practicality of the proposed method for real-world deployment, where rapid response to emerging fault conditions is critical. Overall, the proposed approach provides a scalable and adaptive solution for condition monitoring in UMW and demonstrates strong potential for application to other manufacturing processes characterized by evolving operating conditions and limited labeled data.

Future work will focus on extending the proposed approach to more challenging scenarios involving multiple simultaneous unknown fault types and more severe data imbalance, which are common in large-scale industrial deployments. Another important direction is the adaptation of the proposed method to distributed manufacturing environments, where process data are generated across multiple machines or production lines and cannot be centrally shared due to privacy or proprietary constraints~\cite{eslaminia2024federated}. Enabling effective unknown fault detection and continual learning in such decentralized settings would significantly enhance the practical applicability of the approach. Finally, the current labeling procedure relies on clustering followed by manual inspection. Future work can explore active learning strategies to further reduce labeling effort by prioritizing the most informative samples for human annotation, thereby accelerating model updates under data-scarce conditions.